\documentclass[letterpaper]{article} 
\usepackage[submission]{aaai23}  
\usepackage{times}  
\usepackage{helvet}  
\usepackage{courier}  
\usepackage[hyphens]{url}  
\usepackage{graphicx} 
\usepackage{natbib}  
\usepackage{caption} 
\usepackage{algorithm}
\usepackage{amsmath}
\usepackage{amssymb}
\usepackage{xcolor}
\usepackage{subcaption}
\usepackage{diagbox}
\usepackage{newfloat}
\usepackage{listings}
\DeclareCaptionStyle{ruled}{labelfont=normalfont,labelsep=colon,strut=off} 
\floatstyle{ruled}
\newfloat{listing}{tb}{lst}{}
\floatname{listing}{Listing}
\usepackage{color}
\usepackage{enumitem}
\usepackage{adjustbox}
\usepackage{subcaption}
\usepackage{amsmath}
\usepackage{amssymb}
\usepackage{booktabs}
\usepackage{algorithm}
\usepackage{algpseudocode}
\usepackage{amsmath}
\usepackage{amssymb}
\usepackage{booktabs}
\usepackage{graphicx}
\usepackage{multirow}
\usepackage{subcaption}
\usepackage{bbm}

\title{Pruning On-the-Fly: A Recoverable Pruning Method without Fine-tuning}
\author {
    Dan Liu, 
    Xue Liu 
}
\affiliations {
     McGill University\\
    daniel.liu@mail.mcgill.ca, xue.liu@mcgill.ca
}

\begin{document}

\maketitle

\begin{abstract} Most existing pruning works are resource intensive, as they require retraining or fine-tuning of the pruned models for the purpose of accuracy. We propose a retraining-free pruning method based on hyperspherical learning and loss penalty terms. The proposed loss penalty term pushes some of the model weights far away from zero, while the rest weight values are pushed near zero and can be safely pruned with no need of retraining and a negligible accuracy drop. In addition, our proposed method can instantly recover the accuracy of a pruned model by replacing the pruned values with their mean value. Our method obtains state-of-the-art results in terms of retraining-free pruning and is evaluated on ResNet-18/50 and MobileNetV2 with ImageNet dataset. One can easily get a 50\% pruned ResNet18 model with a 0.47\% accuracy drop. If with fine-tuning, the experiment results show that our method can significantly boost the accuracy of the pruned models compared with existing works. For example, the accuracy of a 70\% pruned (except the first convolutional layer) MobileNetV2 model only drops 3.5\%, much less than the 7\% $\sim$ 10\% accuracy drop with conventional methods.
\end{abstract}

\section{Introduction}
Deep neural network (DNN) models contain millions of parameters, making them impossible to deploy on edge devices. Model size and inference efficiency are major concerns when deploying under resources constraints. Significant research efforts have been made to compress DNN models. Quantization and pruning are popular as they can reduce the model size and computational overhead. 

There are two interesting research topics in pruning: how to reduce the fine-tuning time and how to rapidly recover the network's accuracy from pruning. The purpose of model pruning is to get a DNN model with maximum accuracy and compression ratio. Finding a proper pruning strategy is the main challenge. Most of the existing works need fine-tuning. The steps of pruning and fine-tuning are repeated multiple times to gradually reduce the model size and maintain a higher accuracy. The fine-tuning process is time consuming and requires the whole training dataset. Therefore, studies have been made to explore ways to improve the fine-tuning efficiency and the recovery ability of neural networks with only a few training data. However, some of the weight values are fixated to zero permanently during pruning. The neural network is changing during training, fixing some weight values to zero may restrict its learning ability. The incorrectly pruned weight values are inevitable and hard to be recovered or corrected because the original weight information is lost. Therefore, some researchers propose pruning before or during training so that the network can adapt to pruning. 

In this work, we aim to eliminate the fine-tuning after pruning, i.e., leveraging fine-tuning before pruning to formulate reliable and accurate potential pruning candidates. More specifically, compared with other works that prune the dense model directly, our method reduces the cosine distance between the dense weight values and its pruning mask before the pruning action. Our method is less prone to false pruning as the pruning mask is constantly adapting to the training and the potential pruned weights are pushed near zero. Once the cosine distance is small enough, the model can be pruned to many different sparse levels without any fine-tuning. With our method, a pruned ResNet-18 model can reach up to 50\% sparsity with less than 0.5\% accuracy drop. Combined with our proposed instant recovery method, this sparsity can be pushed up to 70\% with 0.3\% accuracy drop. Our main contributions are as follows:
\begin{itemize}
\item We propose a on-the-fly pruning method which uses regularization terms to minimize the cosine distance between weight values and its pruning mask during training. Once the training is completed, the weight values will be separated into two groups, one being close to zero and the other being far from zero. The processed model can be pruned instantly without any fine-tuning. 

\item We propose a method to increase the model's recovery ability. We show that replacing part of the pruned weights with their mean values can recover part of the model's performance immediately. Our pruning method can greatly improve this instant recovery ability. In addition, our method can significantly improve the pruning potential under high sparsity settings with fine-tuning. For example, for the MobileNetV2 structure with 70\% sparsity (except for the first convolutional layer), the fine-tuned accuracy of other methods drops by 7-10\%, while ours only drops by 3.5\%.

\end{itemize}

\begin{figure*}
     \centering
     \begin{subfigure}[b]{0.22\textwidth}
         \centering
         \includegraphics[width=\textwidth]{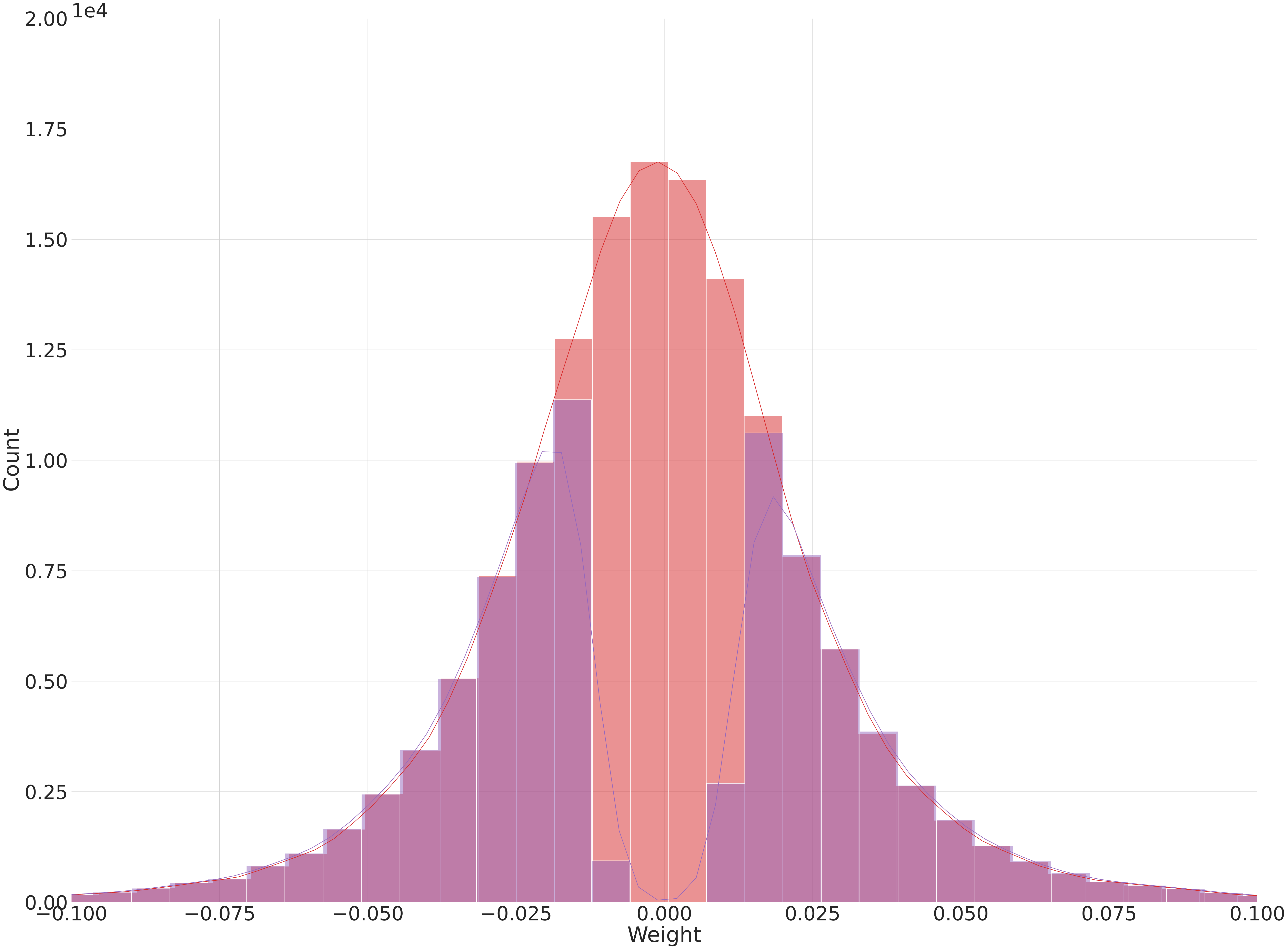}
         \caption{$tr$=0}
         \label{fig:y equals x}
     \end{subfigure}
     \begin{subfigure}[b]{0.22\textwidth}
         \centering
         \includegraphics[width=\textwidth]{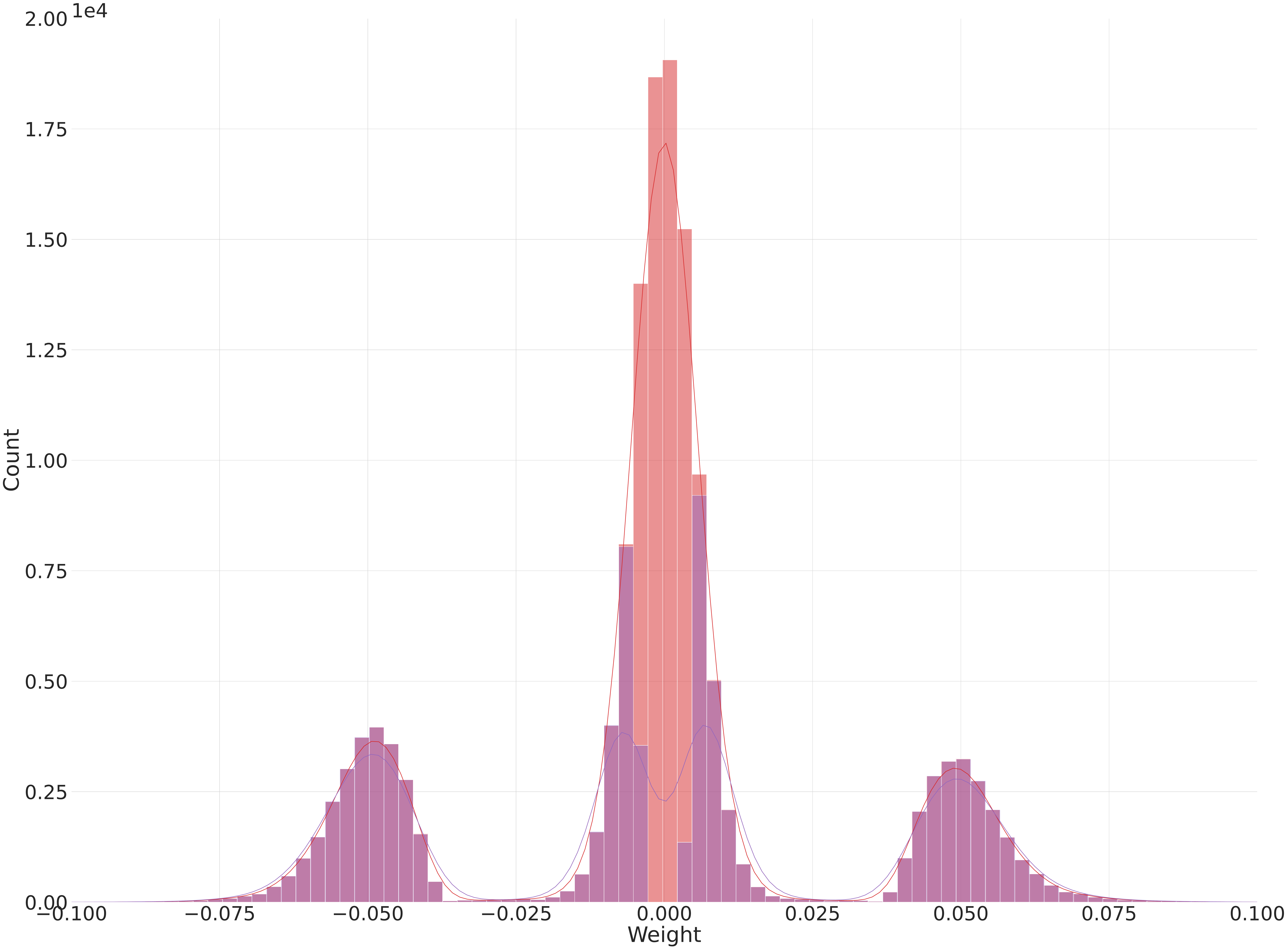}
         \caption{$tr$=0.9}
         \label{fig:three sin x}
     \end{subfigure}
     \begin{subfigure}[b]{0.22\textwidth}
         \centering
         \includegraphics[width=\textwidth]{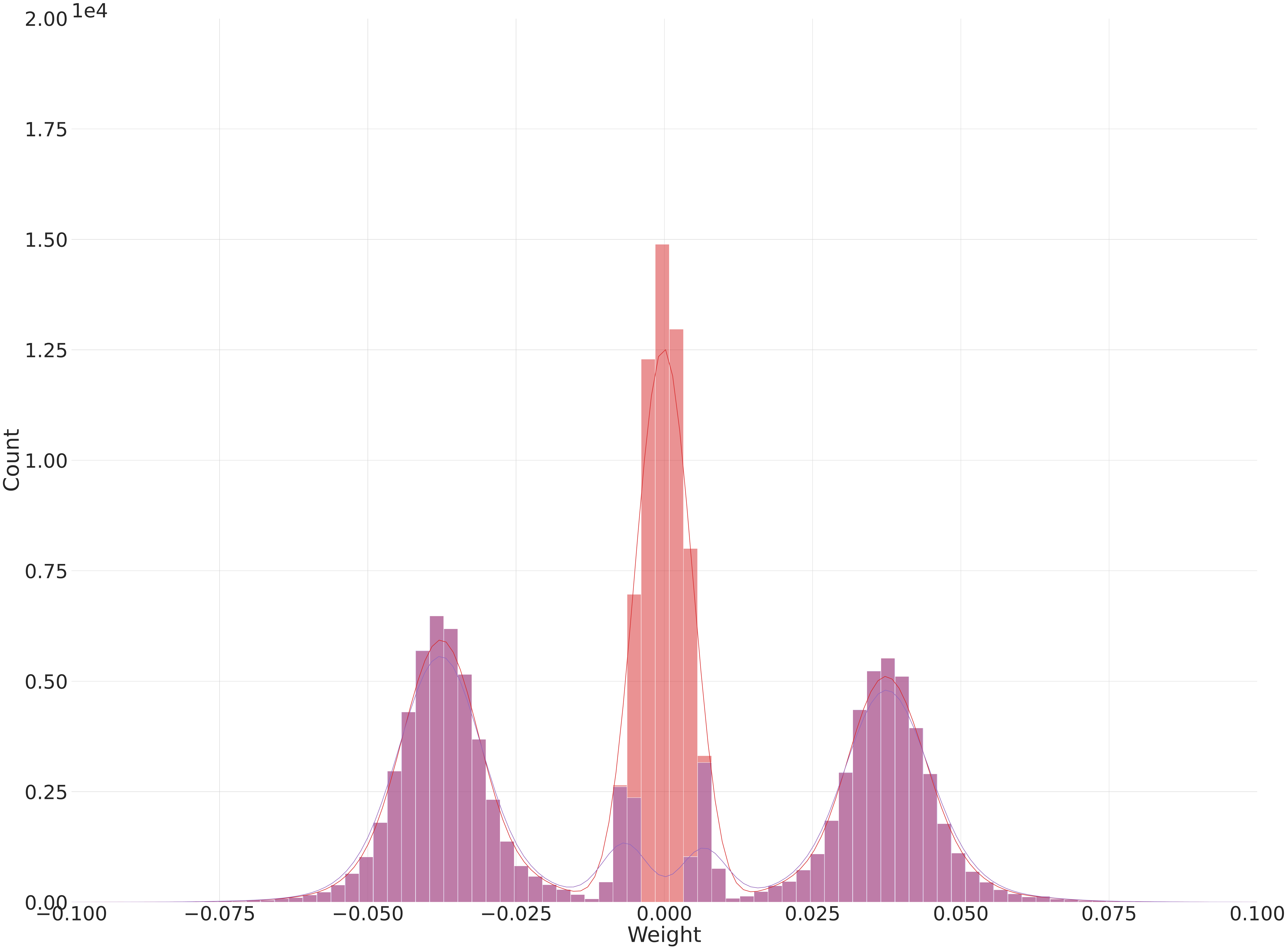}
         \caption{$tr$=0.7}
         \label{fig:three sin x}
     \end{subfigure}
     \begin{subfigure}[b]{0.22\textwidth}
         \centering
         \includegraphics[width=\textwidth]{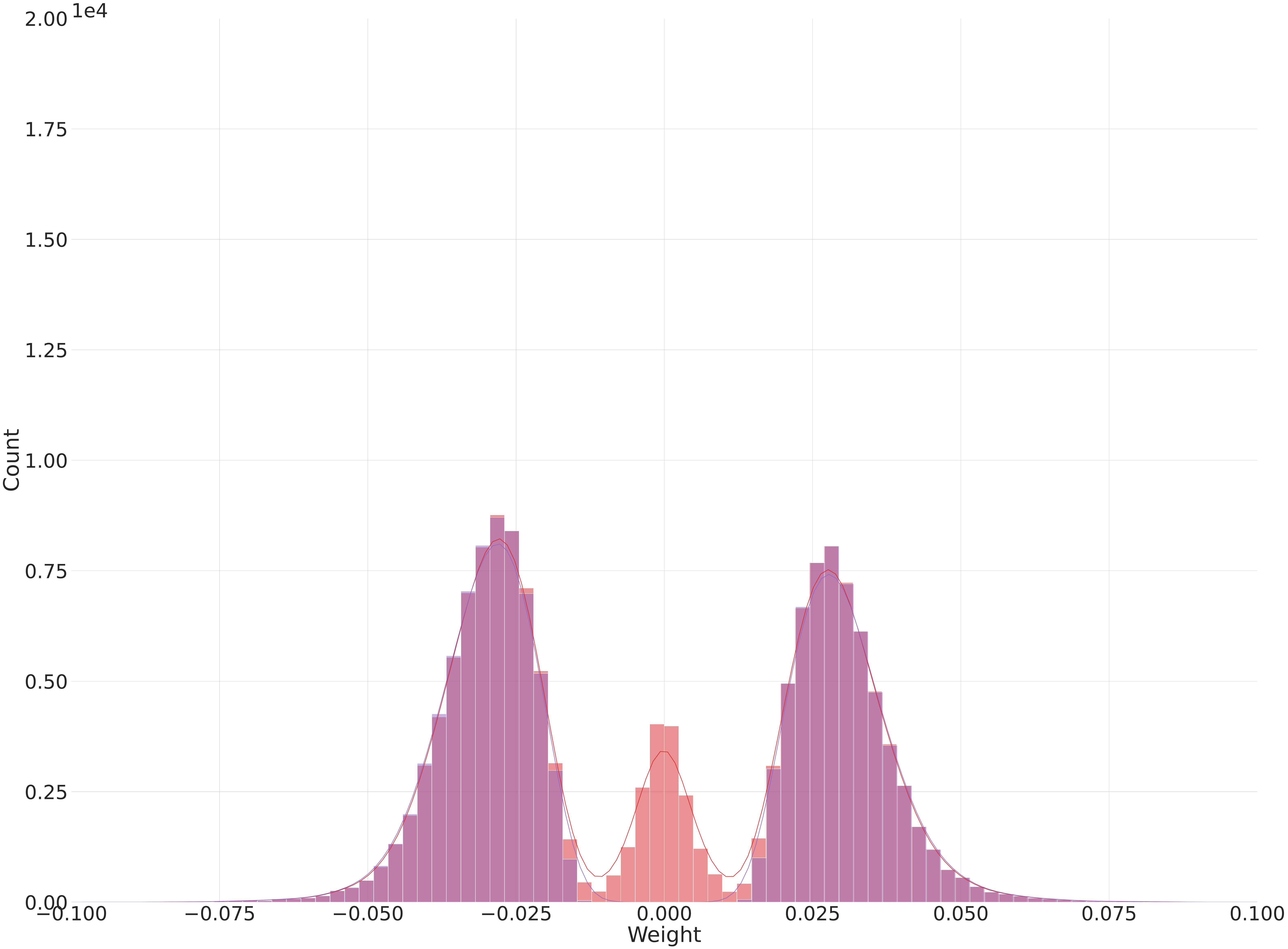}
         \caption{$tr$=0.5}
         \label{fig:three sin x}
     \end{subfigure}
    \begin{subfigure}[b]{0.22\textwidth}
         \centering
         \includegraphics[width=\textwidth]{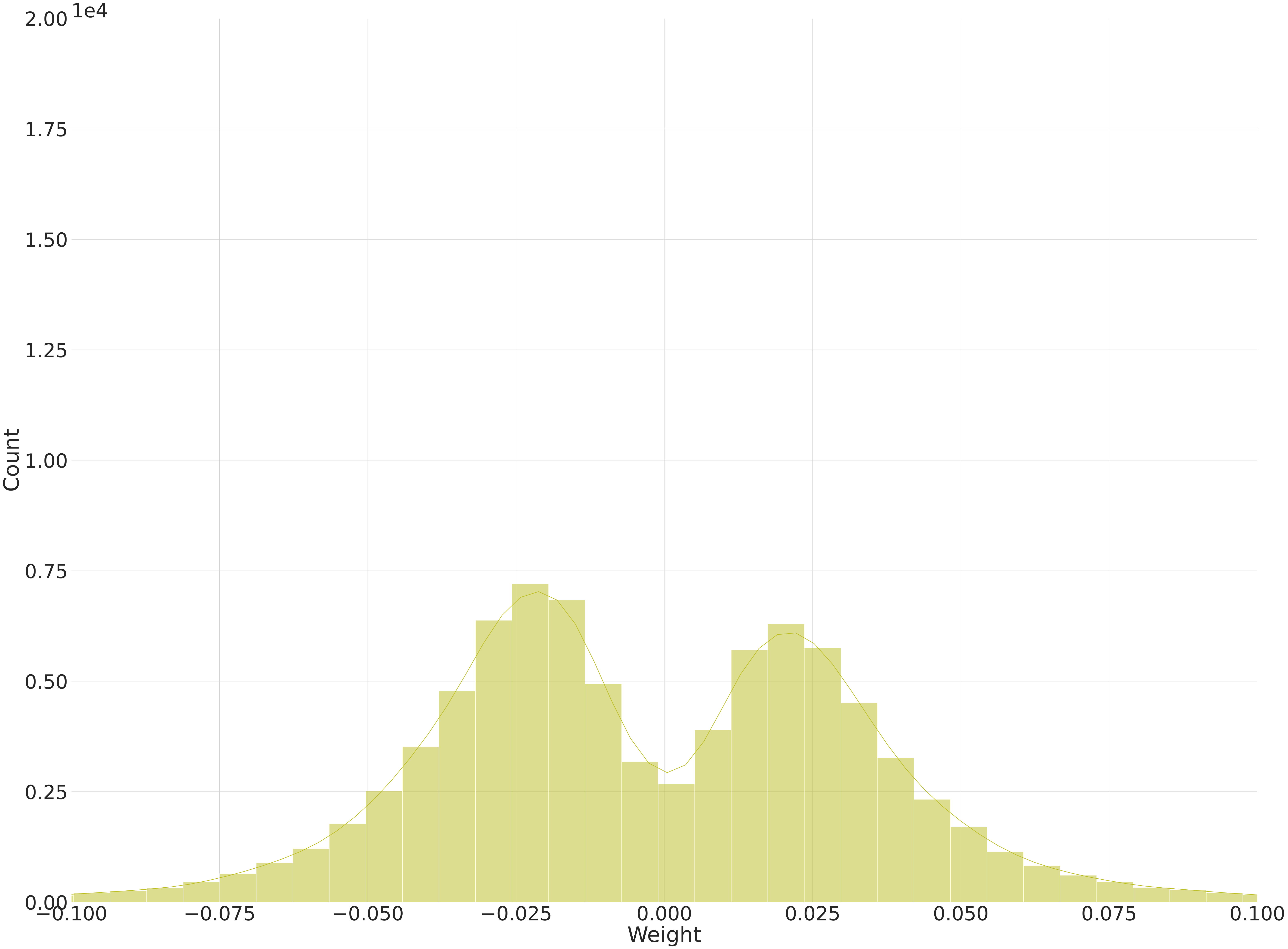}
         \caption{$tr$=0}
         \label{fig:y equals x}
     \end{subfigure}
     \begin{subfigure}[b]{0.22\textwidth}
         \centering
         \includegraphics[width=\textwidth]{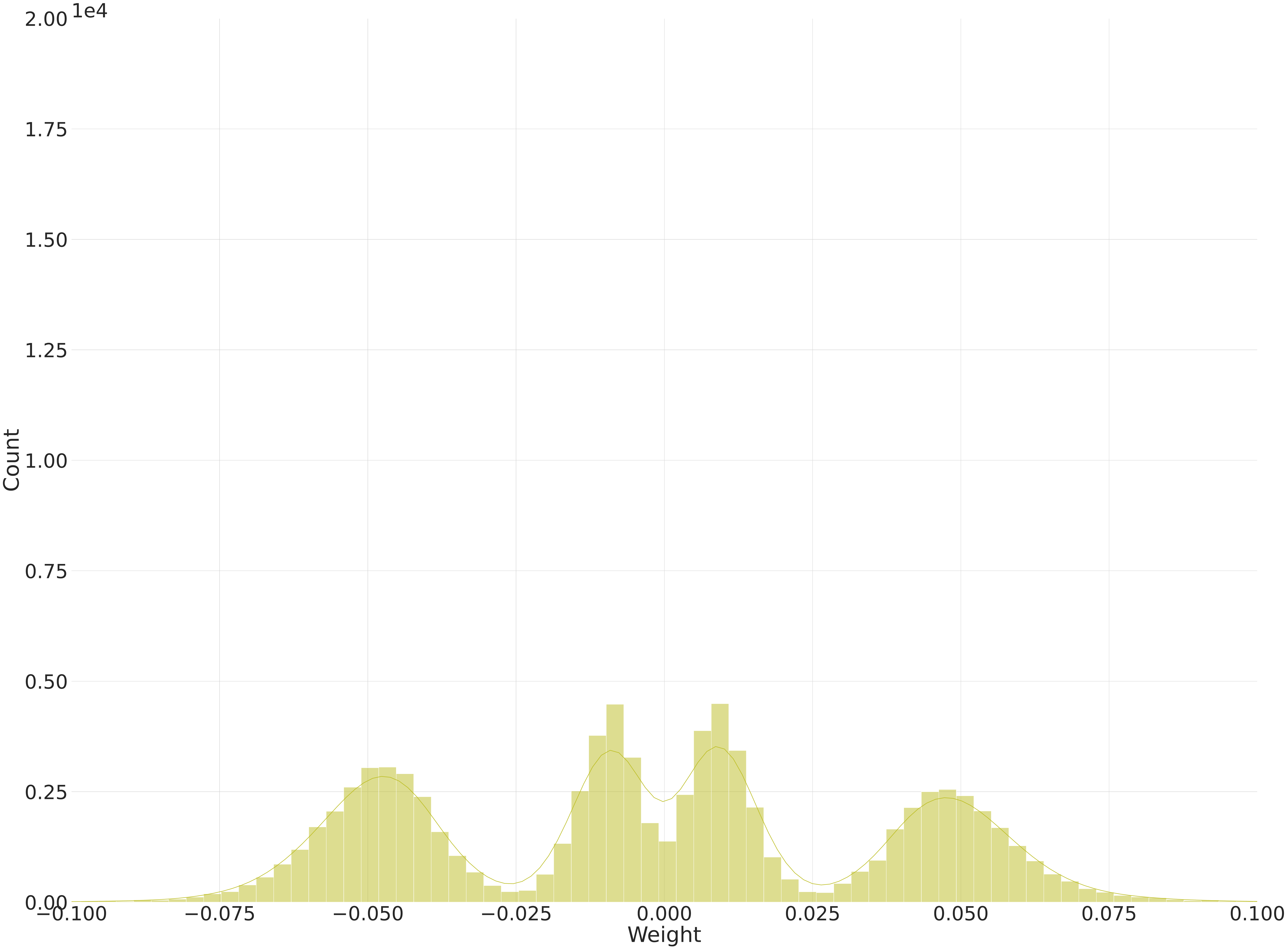}
         \caption{$tr$=0.9}
         \label{fig:three sin x}
     \end{subfigure}
     \begin{subfigure}[b]{0.22\textwidth}
         \centering
         \includegraphics[width=\textwidth]{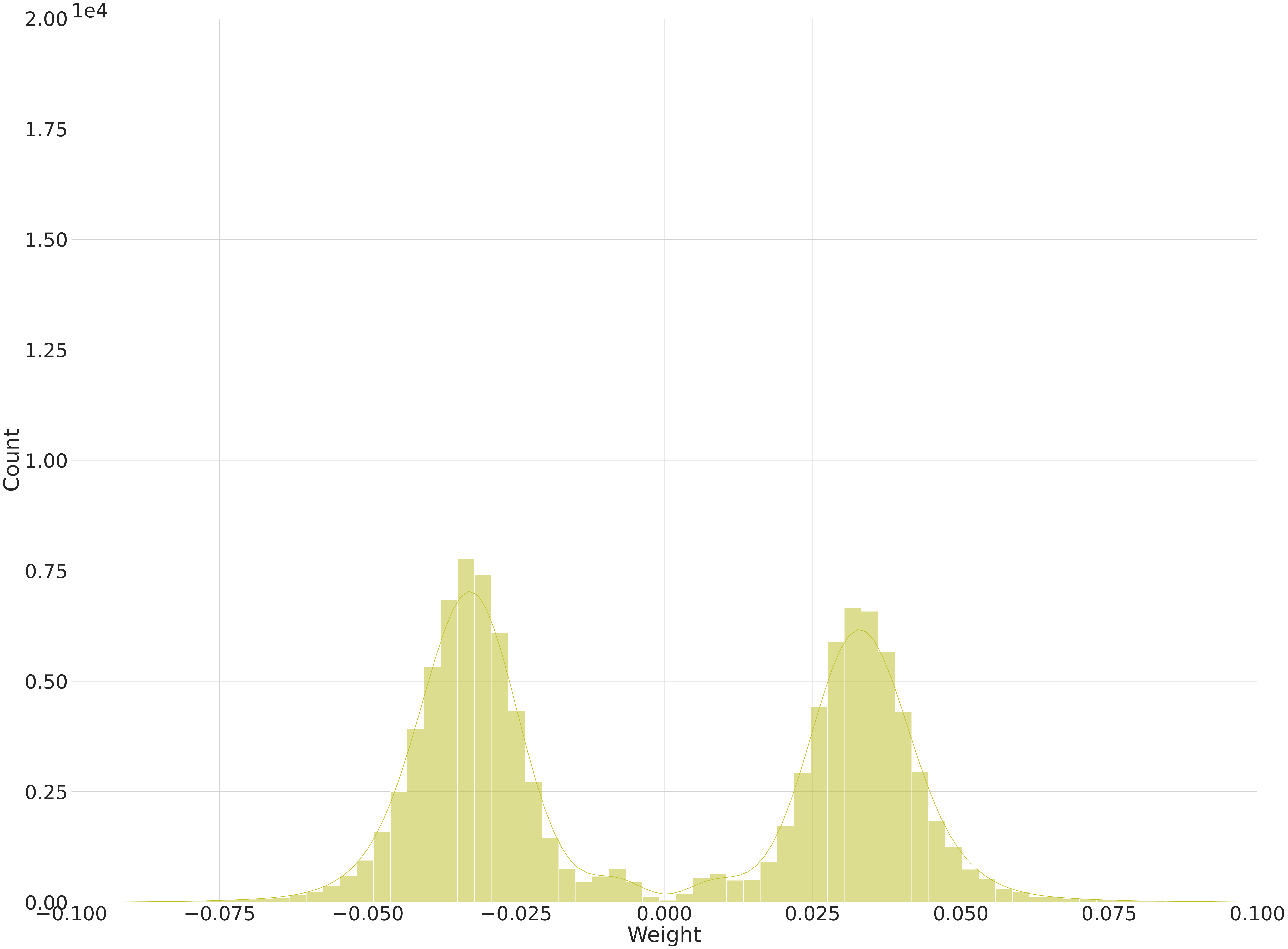}
         \caption{$tr$=0.7}
         \label{fig:three sin x}
     \end{subfigure}
     \begin{subfigure}[b]{0.22\textwidth}
         \centering
         \includegraphics[width=\textwidth]{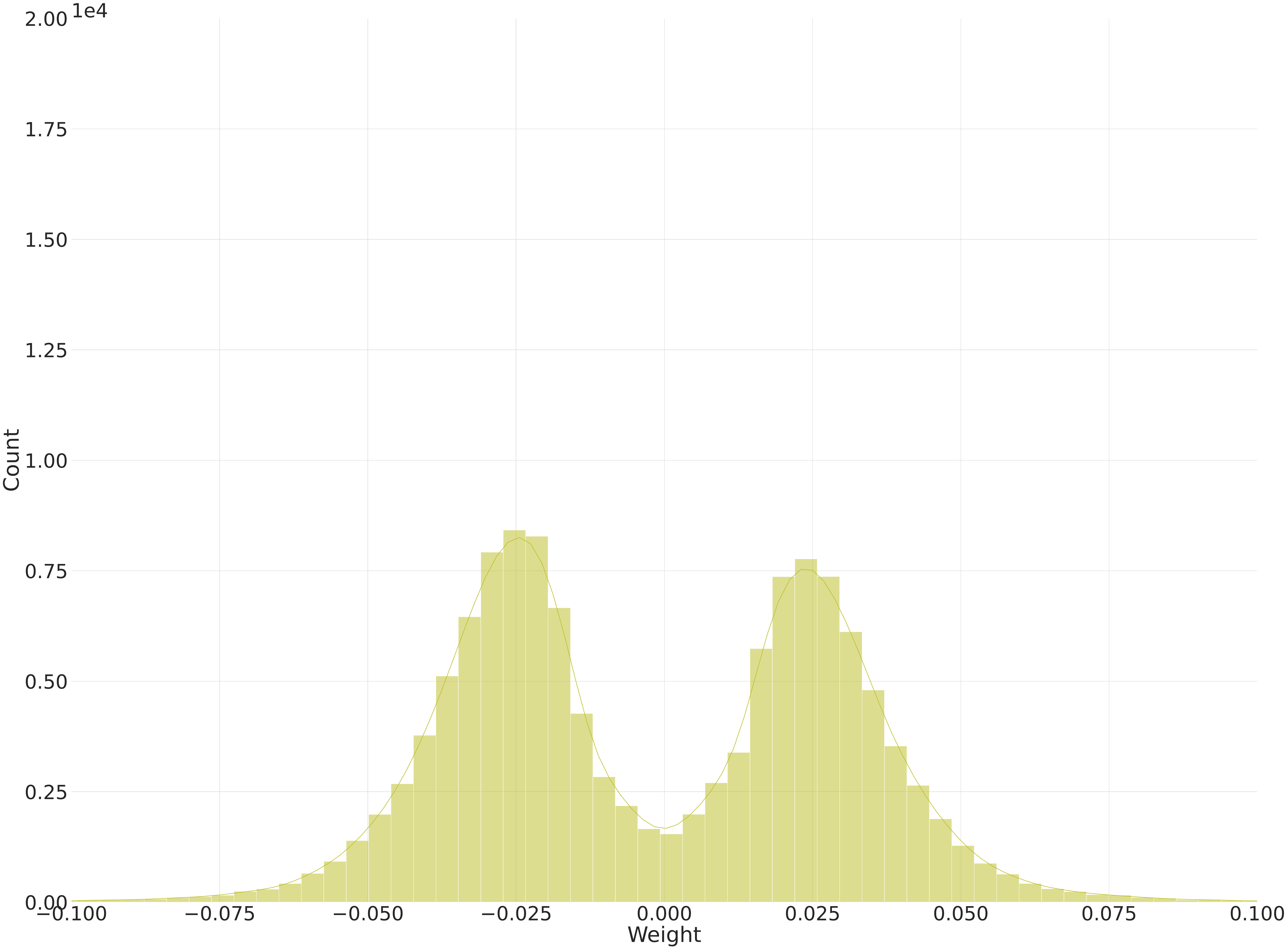}
         \caption{$tr$=0.5}
         \label{fig:three sin x}
     \end{subfigure}
    \caption{\textcolor{black}{The weight distribution of a layer of the baseline models (a), on-the-fly pruning models (b, c, d) and their fine-tuned versions (e, f, g, h). The red columns are the pruned weight values ($r$=0.6, Eq.~\eqref{eq:prune}). The purple columns are the remaining weight values without fine-tuning. The green columns are the fine-tuned weight values. The weight values of (b, c, d) are separated into three parts by the regularization term $L_{tr}$ before pruning and close to their fine-tuned versions (f, g, h). The $tr$ controls the portions of near-zero values (red areas). All of the models in this figure use ResNet-18 structure and the baseline is obtained from PyTorch Zoo.}}
    \label{img:weight_dist}
\end{figure*}
\section{Related Work}
Model compression techniques, such as quantization \cite{wu2016quantized,li2016twn} and pruning \cite{han2015deep,li2016pruning}, have been a trending research topic as they contribute to smaller model sizes and faster inference. A comprehensive overview of model pruning can be found in \cite{liang2021pruning,blalock2020prune_survey}.  
\subsubsection{Pruning Approaches}
Pruning can be categorised by structure and scheduling. The pruning structure specifies whether to prune the layer \cite{chin2018layer,dong2017learning}, the entire kernel \cite{li2016pruning,hu2016network,alvarez2017compression}, or particular weight values \cite{han2015deep}.  Pruning scheduling determines the percentages of the weight values to be removed through each phase. Some techniques perform a one-step pruning to the target weights \cite{liu2018rethinking}. Others change the pruning ratio during training \cite{han2016dsd, gale2019state} or iteratively prune a fixed portion of the weight values across a number of iterations \cite{han2015learning}.  
\subsubsection{Pruning Without Fine-tuning}
Pruning decreases the model's accuracy. Fine-tuning, despite being time consuming, is frequently performed to recover accuracy. Many works explore the possibilities of pruning during or even prior to training without fine-tuning \cite{guo2016dynamic,molchanov2017variational,lee2018snip,gale2019state}. \citeauthor{guo2016dynamic} (\citeyear{guo2016dynamic}) propose a recoverable pruning method using binary mask matrices with a score function to determine whether a single weight value is to be pruned or not. Soft Filter Pruning (SFP) \cite{he2018soft} expands recoverable pruning even further by enabling update of the pruned filters. Sparse variational dropout \cite{molchanov2017variational} employs a dropout hyperparameter that encourages sparsity and acts as a basis for scoring to determine which weights to prune.
Dynamic pruning \cite{lin2017runtime,wu2018blockdrop}, which chooses the pruned weight values using decision components, is another field of research in fine-tuning-free pruning. However, some of the dynamic pruning works are resource intensive as the pruning decision is made in real-time \cite{leroux2017cascading,li2019improved,gao2018dynamic}.

In this work, we study the instant pruning and recovery ability of the neural networks. Compared to the methods stated above, our method aims at using loss penalty to reduce the cosine distance between the dense weight values and its binary mask. Therefore, the optimization process can change the weight distribution(Figure \ref{img:weight_dist}) and benefit pruning. Our method has an advantage over the existing pruning approaches in that it prevents incorrectly pruning and restores the accuracy of the model. Moreover, our strategy does not require a complicated training schedule or sophisticated score function. A simple magnitude-base one-time pruning can produce remarkable outcomes.



\section{Pruning, Recovery and Fine-tuning}


In this section, we show how to perform on-the-fly pruning, and how to recover the model's accuracy instantly or slowly. Before pruning, we apply a regularization term to the objective function to manipulate the weight distribution (Figure \ref{img:weight_dist}). Then the magnitude-based, unstructured pruning can be performed with a negligible accuracy drop.
The pruned model can be recovered by replacing the pruned values with their mean values immediately or recovered by conventional fine-tuning with much higher sparsity. 


\subsection{Regularized Hyperspherical Learning}
Hyperspherical learning  \cite{liu2017deephyperspherical}  restricts the magnitude of the input and weight vectors to one. A general representation of a hyperspherical layer is defined as:
\begin{equation}
\label{eq:spop}
    \mathbf{y}=\phi(\mathbf{W}^\top\mathbf{x}),
\end{equation}
where $\phi$ denotes a nonlinear
activation function, $\mathbf{W} \in \mathbb{R}^{m\times{n}}$ is the weight matrix, $\mathbf{x}\in \mathbb{R}^{m}$ is the input vector to the layer, and $\mathbf{y}\in \mathbb{R}^{n}$ is the output vector.
Each column weight vector $\mathbf{w}_j\in\mathbb{R}^{m}$ of $\mathbf{W}$
subjects to $\|\mathbf{w}_j\|_2=1$ for all $j = 1, . . . , n$, and the input vector $\mathbf{x}$ satisfies $\|\mathbf{x}\|_2=1$.

Given a regular objective function $L$, we formulate the optimization process as:
\begin{equation}
    \label{eq:loss}
     \min_{\mathbf{W}} J(\mathbf{W})=L(\mathbf{W})+ \lambda L_{tr}(\mathbf{W},tr)
\end{equation}
\begin{equation*}
    s.t.~~\mathbf{W}\in\mathbb{R},~0<tr<1.
\end{equation*}
The regularization term $L_{tr}$ is defined as: 
\begin{equation}
    \label{eq:ltr}
    L_{tr}(\mathbf{W},r)=\frac{1}{n}\left(\texttt{trace}(\mathbf{W}^\top{\mathbf{M}}-\mathbf{I})\right)^2 
\end{equation}
\begin{equation*}
    s.t.~~\mathbf{M}=\texttt{HyperSign}(\texttt{Prune}(\mathbf{W},tr)),
\end{equation*}
where $\texttt{Prune}(\cdot)$ denotes magnitude-based unstructured pruning with sparsity $tr$, $\mathbf{M}$ is a mask, and $\texttt{trace}(\cdot)$ returns the trace of a matrix. The quadratic term and $\lambda$ are applied to keep $L(\mathbf{W})$ and $L_{tr}(\mathbf{W},r)$ at the same scale. 

With hyperspherical learning, $\texttt{HyperSign}(\cdot)$ returns the normalized $\texttt{Sign}(\cdot)$, i.e., a pruning mask $\mathbf{M}$ on the hypersphere. More specifically, $m_{ij}\in \{0, \frac{1}{\sqrt{\|\mathbf{m_j}\|_1}}\}$, where $\|\mathbf{m_j}\|_1$ denotes the number of non-zero elements in the $j$-th column vector $\mathbf{m}_j$ of $\mathbf{M}$, and $\|\mathbf{m}_j\|_2=1$ for all $j = 1, . . . , n$. For example, $tr=0.9$ indicates that 90\% of $\mathbf{M}$ is zero.

The diagonal elements (trace) of $\mathbf{W}^\top{\mathbf{M}}$ in Eq.~\eqref{eq:ltr} denotes the cosine similarities between $\mathbf{w_j}$ and $\mathbf{{m}_j}$. Minimizing $L_{tr}$ (Eq.~\eqref{eq:ltr}) is equivalent to pushing $\mathbf{w}_j$ close to $\mathbf{m}_j$, namely, making part of the magnitude of weight values close to $\frac{1}{\sqrt{\|\mathbf{m_j}\|_1}}$ while the rest to zero. Adjusting $tr$ will change the shape of the weight distributions (Figure \ref{img:weight_dist}). The regularization term $L_{tr}$ can be applied to pre-trained and train-from-scratch models. In practice, gradually decreasing $tr$ from a higher value, e.g., $0.9$, to target value, e.g., $0.7$, performs better than using a fixed $tr=0.7$. We compare the mentioned settings in the experiment section.

\subsection{Pruning and Recovery}
\subsubsection{On-the-fly Pruning}
Having adjusting the weight distribution via Eq.~\eqref{eq:ltr}, 
we can prune the model by the magnitude-based unstructured pruning $\texttt{Prune}(\cdot)$, and: 
\begin{equation}
    \label{eq:prune}
    \mathbf{W}=\mathbf{W}'+ \texttt{Prune}(\mathbf{W},r)
\end{equation}
where $0<r<1$ is the pruning ratio, $\mathbf{W}'$ denotes the pruned weight values, and $\texttt{Prune}(\cdot)$ returns the remaining weights. 

Regarding hyperspherical learning, given $\mathbf{W}$, removing its near-zero values $\mathbf{W}'$ or re-scaling the magnitude of $\mathbf{w}_j$ will not affect the model's performance as long as the directions of $\mathbf{W}$ remain unchanged. This allows for on-the-fly pruning with no more fine-tuning afterwards, which is not feasible with conventional pruning methods due to the impact of weight magnitude changes. However, based on our experiment results and other works \cite{lazarevich2021post}, one of the limitations of pruning without fine-tuning is that the accuracy still drops significantly as soon as the pruning ratio exceeds 50\%. We explore two ways to address this issue.
\subsubsection{Instant Recovery with Mean Values}
To recover a pruned model with higher sparsity, we replace the pruned weights $\mathbf{W}'$ with a replacement $\alpha \times \texttt{sign}(\mathbf{W}')$ and minimize the Euclidian distance between them along with a layer-wise scaling factor $\alpha$:
\begin{equation}
    \min_{\alpha} \|\mathbf{W}'-\alpha\mathbf{W}'_{\texttt{sgn}}\|^2_2,
\end{equation}
\begin{equation*}
    s.t.~~\mathbf{W}'_{\texttt{sgn}}\in \{-1,1\}, \alpha>0.
\end{equation*}
And $\alpha$ has a closed form solution \cite{li2016twn}:
\begin{equation}
\label{eq:fill}
    \alpha=\frac{1}{\|\mathbf{W}_{\texttt{sgn}}\|_1}\sum|w'|,
\end{equation} which is the mean magnitude value of the pruned weights. The $\mathbf{W}'_{\texttt{sgn}}$ is important as it keeps the direction information. The work of \cite{zhou2019deconstructing} also supports this intuition by pointing out that the sign is crucial to recover a pruned model. Compared with filling zero, filling the pruned weights with $\alpha\mathbf{W}'_{\texttt{sgn}}$ provides a more accurate approximation of the direction information of $\mathbf{W}$. It is worth noting that this recovery method only works well when $\mathbf{W}'$'s deviation is minor, as a higher deviation indicates a change in direction. Therefore it is preferable to replace part of rather than all of the pruned weight values. This recovery property is related to the pruning ratio $r$ and the regularization term $L_{tr}$. We perform ablation study on them in the experiment section.

Our experimental results further reveal that the instant recovery property exists in both hyperspherical and non-hyperspherical settings. The regularization term $L_{tr} $with hyperspherical learning can stabilize this recovery ability as it produces more near-zero weight values and is less sensitive to weight magnitudes changes.

The instant recovery ability is promising since we only need to store a binary mask $\mathbf{W}'_{\texttt{sgn}}$ and a mean value $\alpha$ for each layer to compress models. With the proposed instant recovery method, a hyperspherical ResNet-18 model can be pruned up to 70\% sparsity with 0.3\% accuracy drop without any fine-tuning, which outperforms most of the existing fine-tuning-based pruning works. However, in practice, we observe that the instant recovery only works well on MobileNetV2 with sparsity less than 30\%. 

\subsubsection{Slow Recovery with Fine-tuning}
The instant recovery method only benefits model size compression, while it does not reduce FLOPs as the zero values are replaced by mean values. Therefore, we further explore the impact of $L_{tr}$ on fine-tuning-based pruning. We follow the conventional way in to fine-tune the processed models \cite{blalock2020prune_survey}. 

\section{Training Details}
Our proposed method can be applied to from-scratch-training, with more training efforts and slightly lower accuracy though, as the $L_{tr}$ term relies on converged model weights as do conventional pruning methods. All of the models in our experiment are initialized from PyTorch Zoo except the training-from-scratch ones in ablation study. The recommended initial settings for $L_{tr}$ are $tr=0.9$, and $0.5<\lambda<2$. The initial $\lambda$ depends on different network structure. For example, $\lambda=2$ for ResNet18/MobileNetV2, and $\lambda=1$ for ResNet50. The overall process can be summarized as:
i) Fine-tuning with hyperspherical learning \cite{liu2017deephyperspherical} and $L_{tr}$ from pre-trained PyTorch Zoo models with specific $tr$ (Eq.~\eqref{eq:ltr}); ii) Apply instant recovery or fine-tuning after unstructured pruning.

We use PyTorch mixed precision with 8$\times$ Nvidia A100 GPU for training. We use the cosine annealing schedule with restarts (every 10 epochs) \cite{loshchilov2016sgdr} to adjust the learning rates. The initial learning rate is 0.01 and the batch size is 256. Gradually decreasing $tr$ from $0.9$ to $0.7$ within 90 epochs can obtain a good result.

\section{Experiment}
In this section, we study the impact of $tr$ to the accuracy of on-the-fly pruning, instant recovery, and fine-tuning. We perform image classification task to evaluate our proposed method on the ImageNet dataset \cite{ILSVRC15} with ResNet-18/50 \cite{he2016deep} and MobileNetV2 \cite{sandler2018mobilenetv2} architectures. We use ``+'' to denote training from scratch models. The other models are initialized by pre-trained weights provided by the PyTorch zoo.``$\rightarrow$'' denotes gradually decrease. ``$tr$=0'' denotes baseline model obtained from PyTorch.  

\subsection{Experimental Setup}
The batch size is 256. The weight decay is 0.0001, and the momentum of stochastic gradient descent (SGD) is 0.9. During fine-tuning, we use the cosine annealing schedule with restarts \cite{loshchilov2016sgdr} to adjust the learning rates. The initial learning rate is 0.01. When training from scratch, we follow the recipe from PyTorch. We prune all of the linear and convolutional layers except the first convolutional layer.
\subsubsection{Training from Scratch}
Our proposed $L_{tr}$ can be applied directly to training-from-scratch. Table \ref{R18_fly} shows the comparison results of pre-trained and trained-from-scratch method. As we stated above, the on-the-fly pruning accuracy of the training-from-scratch model is slightly worse than the pre-trained ones. In the Table \ref{R18_recover}, we also compare the recovery ability between them. Although the training-from-scratch models are not as good as the pre-trained ones, they still can outperform the baseline model. In addition, when training from scratch, the fixed $tr=0.95$ performs better than the gradually decreased ones, i.e. $tr=0.9\rightarrow0.7$.

\subsection{Pruning On-the-fly}
Pruning on-the-fly means pruning directly by using unstructured \cite{han2015deep} method. We compare different settings of sparsity and $tr$ to study their impact to the performance. The results are shown in Table \ref{R18_fly}, \ref{R50_fly} and \ref{Mob_fly}. 

The impact of different $tr$ and sparsity is listed in the Table \ref{R18_fly}. The pre-trained models with gradually decreased $tr$ significantly outperform the baseline models. We also compare the impact on ResNet-50 (Table \ref{R50_fly}), and MobileNetV2 (Table \ref{Mob_fly}).

\begin{table}[!ht]
\centering
        \begin{adjustbox}{max width=0.5\textwidth}
    \begin{tabular}{l|l|ll|cc|ll}
        \toprule
       
       \textbf{Sparsity}& $tr$=$0$ & $tr$=$0.95$ & $tr$=${0.9}$ & $tr$=${0.95\rightarrow0.9}$ & $tr$=${0.9\rightarrow0.7}$ & $tr$=${0.9\rightarrow0.7}^+$ & $tr$=${0.9}^+$ \\ 
        \midrule
        Dense & 69.54 & 68.80 & 69.74 & 69.47 & 69.74 & 69.03 & 69.19  \\ 
        \midrule
        30\% & 68.98 & 68.50 & 69.41 & 69.31 & \textbf{69.55} & 68.61 & 68.78 \\ 
        50\% & 65.20 & 66.65 & 60.00& 68.14 & \textbf{68.97}  & 67.13 & 66.79 \\ 
        70\% & 40.03 & 53.44 & 9.70 & 62.99 & \textbf{64.82} & 24.08 & 54.77 \\ 
    \bottomrule
        \end{tabular}
        \end{adjustbox}
    \caption{The instant pruning accuracy of ResNet18 on the ImageNet dataset. ``+'' denotes training from scratch. ``$\rightarrow$'' denotes gradually decrease. ``$tr$=0'' denotes baseline model from PyTorch.}
    \label{R18_fly}
\end{table}
\begin{table}[!ht]
    \footnotesize
        \centering
        \begin{adjustbox}{max width=0.5\textwidth}
    \begin{tabular}{llc}
        \toprule
        \textbf{Sparsity} & \textbf{Baseline}$_{tr=0}$ & $_{tr=0.9\rightarrow0.7}$ \\ 
        \midrule
        Dense & 76.15 & 77.15\\ 
        \midrule
        30\% & 75.28 & 76.94 \\ 
        50\% & 72.90 & 76.62 \\ 
        70\% & 43.82 & 69.01 \\ 
    \bottomrule
        \end{tabular}
        \end{adjustbox}
    \caption{The instant pruning accuracy of ResNet-50 on the ImageNet dataset. }
    \label{R50_fly}
\end{table}
\begin{table}[!ht]
    \footnotesize
        \centering
        \begin{adjustbox}{max width=0.5\textwidth}
    \begin{tabular}{lll}
    \toprule
       \textbf{Sparsity} & \textbf{Baseline}$_{tr=0}$ & $_{tr=0.9\rightarrow0.7}$ \\ 
        \midrule
        Dense & 71.35 & 70.75 \\ 
        \midrule
        10\% & 71.28 & 70.59 \\ 
        20\% & 69.50 & 68.57 \\ 
        30\% & 59.12 & 60.60 \\ 
        40\% & 13.80 & 34.85 \\ 
    \bottomrule
        \end{tabular}
        \end{adjustbox}
    \caption{The instant pruning accuracy of MobileNetV2 on the ImageNet dataset. }
    \label{Mob_fly}
\end{table}

\subsection{Instant Recovery}
In this section, we compare the instant recovery ability. We fill part of the pruned weight values. ``\textbf{Sp}$_1$'' denotes the starting point and ``\textbf{Sp}$_2$'' denotes the target sparsity. For example, ``\textbf{Sp}$_1$=0.3'' and ``\textbf{Sp}$_2$=0.5'' means the pruned weight values in the pruning range from 30\% to 50\% are replaced by Eq.~\eqref{eq:fill}, i.e., $\mathbf{W}'=\texttt{Prune}(\mathbf{W},0.3)-\texttt{Prune}(\mathbf{W},0.5)$; the sparsity is 50\% and with one extra mask $\mathbf{W}'$.

We compare with baseline models from PyTorch, DPF \cite{lin2020dynamic_DPF}, and DSR \cite{mostafa2019parameter}. Unlike other fine-tuning-free methods, our method can produce models with different sparsity and accuracy. We also observe that once the sparsity is close to $tr$, the instant recovery ability drops quickly (Table \ref{R50_recover}). The MobileNetV2 is very sensitive to the instant pruning (Table \ref{Mob_recover}) and the performance is not as good as the baseline models.
\begin{table}[h!]
     \footnotesize
        \centering
        \begin{adjustbox}{max width=0.5\textwidth}
    \begin{tabular}{ll|l|ll|cc|ll}
     \toprule
        \textbf{Sp}$_1$ & \textbf{Sp}$_2$ &$tr$=$0$ & $tr$=$0.9$& $tr$=$0.95$ & $tr$=${0.9\rightarrow0.7}$ & $tr$=${0.95\rightarrow0.9}$ &  $tr$=${0.9\rightarrow0.7}^+$ & $tr$=${0.9}^+$ \\ 
        \midrule
        &Dense & 69.54 & 68.80 & 69.74 & 69.47 & 69.74 & 69.03 & 69.19   \\ 
        \midrule
        10\% & 40\% & 69.41 & 69.71 & 68.81 & 68.75 & \textbf{69.45} & 68.86 & 69.11 \\ 
        ~ & 60\% & 68.67 & 69.46 & 68.42 & 63.69 & \textbf{69.13} & 66.07 & 68.49 \\ 
        ~ & 80\% & 65.15 & 47.28 & 66.38 & 53.62 & \textbf{67.71} & 38.45 & 66.00 \\ 
        \midrule
        30\% & 50\% & 68.98 & 69.54 & 68.51 & 68.99 & \textbf{69.29} & 68.71 & 68.77 \\ 
        ~ & 70\% & 68.32 & 69.07 & 68.29 & 67.51 & \textbf{69.08} & 61.35 & 68.36 \\ 
        ~ & 90\% & 62.78 & 50.07 & \textbf{63.68} & 58.45 & 35.56 & 31.15 & 60.00 \\ 
    \bottomrule
        \end{tabular}
        \end{adjustbox}
    \caption{The instant recovery accuracy of ResNet18 on the ImageNet dataset.``+'' denotes training from scratch. ``$\rightarrow$'' denotes gradually decrease. ``$tr$=0'' denotes baseline model from PyTorch.}
    \label{R18_recover}
\end{table}

\begin{table}[!ht]
    \footnotesize
        \centering
        \begin{adjustbox}{max width=0.5\textwidth}
    \begin{tabular}{llll}
    \toprule
        \textbf{Sp}$_1$ & \textbf{Sp}$_2$  & \textbf{Baseline}$_{tr=0}$ & $_{tr=0.9\rightarrow0.7}$ \\ 
        \midrule
        & Dense & 76.15 & 77.15 
        \\ 
        \midrule
        10\% & 40\% & 75.69 & 77.15 \\ 
        ~ & 60\% & 75.44 & 76.91 \\ 
        ~ & 80\% & 73.09 & 33.16 \\ 
        \midrule
        30\% & 50\% & 75.26 & 76.93 \\ 
        ~ & 70\% & 74.93 & 76.28 \\ 
        ~ & 90\% & 71.43 & 2.36 \\ 
    \bottomrule
        \end{tabular}
        \end{adjustbox}
    \caption{The instant recovery accuracy of ResNet-50 on the ImageNet dataset.}
    \label{R50_recover}
\end{table}

\begin{table}[!ht]
    \footnotesize
        \centering
        \begin{adjustbox}{max width=0.5\textwidth}
    \begin{tabular}{llll}
    \toprule
       \textbf{Sp}$_1$ & \textbf{Sp}$_2$  & \textbf{Baseline}$_{tr=0}$ & $_{tr=0.9\rightarrow0.7}$ \\ 
        \midrule
         & Dense & 71.88 & 70.75 
         \\ 
         \midrule
        10\% & 40\% & 69.47 & 66.73 \\ 
        ~ & 60\% & 59.84 & 6.72 \\ 
        ~ & 80\% & 8.25 & - \\ 
        \midrule
        30\% & 50\% & 58.71 & 55.1 \\ 
        ~ & 70\% & 47.8 & 3.82 \\ 
        ~ & 90\% & 1.99 & - \\ 
    \bottomrule
        \end{tabular}
        \end{adjustbox}
    \caption{The instant recovery accuracy of MobileNetV2 on the ImageNet dataset.}
    \label{Mob_recover}
\end{table}

\begin{table}[h!]
    \footnotesize
        \centering
        \begin{adjustbox}{max width=0.5\textwidth}
        \begin{tabular}{lllll}
        \toprule
        \textbf{Methods}& 
        \textbf{Dense} & 
        \textbf{Sparse}&
        \textbf{Diff.}&
        \textbf{Sparsity}
        \\
        \midrule

         \textsc{DSR \cite{mostafa2019parameter}} & 74.9   & 71.6&3.30&80\%  \\
         \textsc{DPF \cite{lin2020dynamic_DPF}} & 75.95   & 75.13&0.82&80\%  \\
          \textsc{\textbf{POF (Ours)}}$_{tr=0.95}$ &  75.85  & \textbf{75.27} & \textbf{0.58}&35-80\%  \\
         \midrule
         \textsc{DSR \cite{mostafa2019parameter}} & 74.9   & 73.3&1.6&71.4\%  \\
         \textsc{DPF \cite{lin2020dynamic_DPF}} & 75.95   & 75.48&0.47&73.5\%  \\
          \textsc{\textbf{POF (Ours)}}$_{tr=0.9}$ & 75.85 & \textbf{75.51} & \textbf{0.34}&40-70\%  \\ 
        
        \bottomrule
        \end{tabular}
        \end{adjustbox}
    \caption{The instant recovery accuracy of ResNet50 on the ImageNet dataset. ``Diff.'' denotes the accuracy difference between the dense and sparse models. ``35-80\%'' means $\mathbf{W}'=\texttt{Prune}(\mathbf{W},0.35)-\texttt{Prune}(\mathbf{W},0.8)$(Eq.~\eqref{eq:fill}).}
    \label{tab_w2a32}
\end{table} 

\subsection{Fine-tuning}
We compare our result with one-shot pruning \cite{han2015deep}, gradual pruning\cite{zhu2017prune}, and cyclical pruning \cite{srinivas2022cyclical}. The fine-tuning accuracy is outstandingly improved by our proposed method. For example, the MobileNetV2 with 50\% sparsity even outperforms the accuracy of the original dense model (Table \ref{Mob_FT}); with 70\% sparsity, our method only brings 3.54\% accuracy drop, whereas the conventional pruning methods reduce 7\% - 10\% of accuracy.
\begin{table}[h!]
    \footnotesize
        \centering
        \begin{adjustbox}{max width=0.5\textwidth}
        \begin{tabular}{lllll}
        \toprule
        \textbf{Methods}& 
        \textbf{Dense} & 
        \textbf{Sparse}&
        \textbf{Diff.}&
        \textbf{Sparsity}
        \\
        \midrule

          \textsc{One-shot \cite{han2015learning}} & 69.70  & 63.50 &4.2 & 90\% \\
         \textsc{Gradual\cite{zhu2017prune}} & 69.70  & 63.60 & 4.1&90\%\\
         \textsc{Cyclical \cite{srinivas2022cyclical}} & 69.70  & 64.90&4.8&90\%  \\
         \textsc{\textbf{POF (Ours)}}$_{tr=0.95}$ &  {68.86}  & \textbf{65.69} & \textbf{3.17}&90\%  \\
          \textsc{\textbf{POF (Ours)}}$_{tr=0.9}$ & 69.73 & 64.95 & 4.78&90\%  \\
           \textsc{\textbf{POF (Ours)}}$_{tr=0.7}$ & 69.73  & 63.02 & 6.68 &90\%  \\
          \cmidrule{1-5}
        \textsc{One-shot \cite{han2015learning}} & 69.70  & 68.20 &1.5 & 80\% \\
         \textsc{Gradual\cite{zhu2017prune}} & 69.70  & 67.80 & 1.9&80\%\\
         \textsc{Cyclical \cite{srinivas2022cyclical}} & 69.70  & 68.30&1.4&80\%  \\
         \textsc{\textbf{POF (Ours)}}$_{tr=0.95}$ &  {68.86}  & {68.67} & \textbf{0.19}&80\%  \\
          \textsc{\textbf{POF (Ours)}}$_{tr=0.9}$ & 69.73 & \textbf{69.09} & 0.64&80\%  \\
           \textsc{\textbf{POF (Ours)}}$_{tr=0.7}$ & 69.73  & 68.89 & 0.84 &80\%  \\
            \cmidrule{1-5}
        \textsc{One-shot \cite{han2015learning}} & 69.70  & 69.20 &0.5 & 70\% \\
         \textsc{Gradual\cite{zhu2017prune}} & 69.70  & 69.20 & 0.5&70\%\\
         \textsc{Cyclical \cite{srinivas2022cyclical}} & 69.70  & 69.40&0.3&70\%  \\
         \textsc{\textbf{POF (Ours)}}$_{tr=0.95}$ &  {68.86}  & {69.99} & \textbf{-1.13}&70\%  \\
          \textsc{\textbf{POF (Ours)}}$_{tr=0.9}$ & 69.73 & \textbf{70.44} & -0.71&70\%  \\
           \textsc{\textbf{POF (Ours)}}$_{tr=0.7}$ & 69.73  & 69.97 & -0.24 &70\%  \\
           \cmidrule{1-5}
        \textsc{One-shot \cite{han2015learning}} & 69.70  & 69.90 &-0.2 & 60\% \\
         \textsc{Gradual\cite{zhu2017prune}} & 69.70  & 69.90 & -0.2&80\%\\
         \textsc{Cyclical \cite{srinivas2022cyclical}} & 69.70  & 69.60&0.1&60\%  \\
         \textsc{\textbf{POF (Ours)}}$_{tr=0.95}$ &  {68.86}  & {70.42} & \textbf{-1.56}&60\%  \\
          \textsc{\textbf{POF (Ours)}}$_{tr=0.9}$ & 69.73 & \textbf{70.48} & -0.75&60\%  \\
           \textsc{\textbf{POF (Ours)}}$_{tr=0.7}$ & 69.73  & 70.28 & -0.55 &60\%  \\
        
        
       
         
        \bottomrule
        \end{tabular}
        \end{adjustbox}
    \caption{The fine-tuned Top-1 test accuracy of ResNet18 on the ImageNet dataset. Our method (POF) outperforms the existing methods in terms of accuracy and difference (``Diff.''). In addition, our method has a relative lower starting accuracy.}
      \label{R18_FT}
\end{table} 

\begin{table}[h!]
    \footnotesize
        \centering
        \begin{adjustbox}{max width=0.5\textwidth}
        \begin{tabular}{lllll}
        \toprule
        \textbf{Methods}& 
        \textbf{Dense} & 
        \textbf{Sparse}&
        \textbf{Diff.}&
        \textbf{Sparsity}
        \\
        \midrule

          \textsc{One-shot \cite{han2015learning}} & 76.16  & 72.8 &3.36 & 90\% \\
         \textsc{Gradual\cite{zhu2017prune}} & 76.16  & 71.9 & 4.26&90\%\\
         \textsc{Cyclical \cite{srinivas2022cyclical}} & 76.16  & 73.3&2.86&90\%  \\
         \textsc{\textbf{POF (Ours)}}$_{tr=0.95}$ &  {77.00}  & {73.79} & {3.21}&90\%  \\
          \textsc{\textbf{POF (Ours)}}$_{tr=0.9}$ & 76.62 & \textbf{74.68} & \textbf{1.94}&90\%  \\
           \textsc{\textbf{POF (Ours)}}$_{tr=0.7}$ & 77.15  & 5.82 & 71.33 &90\%  \\
          \cmidrule{1-5}
        \textsc{One-shot \cite{han2015learning}} &  76.16  & 75.4 &0.76 & 80\% \\
         \textsc{Gradual\cite{zhu2017prune}} &  76.16    & 74.9 & 1.26&80\%\\
         \textsc{Cyclical \cite{srinivas2022cyclical}} & 76.16   & 75.3&0.86&80\%  \\
         \textsc{\textbf{POF (Ours)}}$_{tr=0.95}$ &  {77.00}  & \textbf{76.24} & \textbf{0.39}&80\%  \\
          \textsc{\textbf{POF (Ours)}}$_{tr=0.9}$ & 76.62 & {76.23} & 0.64&80\%  \\
           \textsc{\textbf{POF (Ours)}}$_{tr=0.7}$ & 77.15  & 76.02 & 1.13 &80\%  \\
            \cmidrule{1-5}
         \textsc{One-shot \cite{han2015learning}} &  76.16  & 75.9 &0.26 & 70\% \\
         \textsc{Gradual\cite{zhu2017prune}} &  76.16    & 75.8 & 0.36&70\%\\
         \textsc{Cyclical \cite{srinivas2022cyclical}} & 76.16   & 75.7&0.46&70\%  \\
         
         \textsc{\textbf{POF (Ours)}}$_{tr=0.95}$ &  {77.00}  & \textbf{76.81} & {0.19}&70\%  \\
          \textsc{\textbf{POF (Ours)}}$_{tr=0.9}$ & 76.62 & {76.5} & \textbf{0.12}&70\%  \\
           \textsc{\textbf{POF (Ours)}}$_{tr=0.7}$ & 77.15  & 76.74 & 0.41 &70\%  \\
        
        
       
         
        \bottomrule
        \end{tabular}
        \end{adjustbox}
    \caption{The fine-tuned Top-1 test accuracy of ResNet50 on the ImageNet dataset. ``POF'' denotes our proposed method. ``Diff.'' denotes the accuracy difference between the dense and sparse models.}
     \label{R50_FT}
\end{table} 
\begin{table}[t]
    \footnotesize
        \centering
        \begin{adjustbox}{max width=0.5\textwidth}
        \begin{tabular}{lllll}
        \toprule
        \textbf{Methods}& 
        \textbf{Dense} & 
        \textbf{Sparse}&
        \textbf{Diff.}&
        \textbf{Sparsity}
        \\
        \midrule

         \textsc{One-shot \cite{han2015learning}} &  71.7  & 61.3 &10.4 & 70\% \\
         \textsc{Gradual\cite{zhu2017prune}} &  71.7    & 62.7 & 9&70\%\\
         \textsc{Cyclical \cite{srinivas2022cyclical}} & 71.7  & 64.4&7.3&70\%  \\
         
          \textsc{\textbf{POF (Ours)}}$_{tr=0.9}$ & 70.43 & \textbf{66.89} & \textbf{3.54}&70\%  \\
         
          \cmidrule{1-5}
        \textsc{One-shot \cite{han2015learning}} &  71.7  & 66.7 &5 & 60\% \\
         \textsc{Gradual\cite{zhu2017prune}} &  71.7   & 67.6 & 4.1&60\%\\
         \textsc{Cyclical \cite{srinivas2022cyclical}} & 71.7   & 68.4&3.3&60\%  \\
          \textsc{\textbf{POF (Ours)}}$_{tr=0.9}$ & 70.43 & \textbf{70.22} & \textbf{0.21}&60\%  \\
            \cmidrule{1-5}
         \textsc{One-shot \cite{han2015learning}} & 71.7  &67.6 &4.1 & 50\% \\
         \textsc{Gradual\cite{zhu2017prune}} & 71.7  & 69.8 & 1.9&50\%\\
         \textsc{Cyclical \cite{srinivas2022cyclical}} & 71.7  & 70.1&1.6&50\%  \\
          \textsc{\textbf{POF (Ours)}}$_{tr=0.9}$ & 70.43 & \textbf{71.77} & \textbf{-1.34}&50\%  \\
        
        
       
         
        \bottomrule
        \end{tabular}
        \end{adjustbox}
    \caption{The fine-tuned Top-1 test accuracy of MobileNetV2 on the ImageNet dataset. Our method (``POF'') significantly outperform the existing results.}
    \label{Mob_FT}
\end{table}

\section{Conclusion}
In this paper, we show that hyperspherical learning with loss regularization term can greatly improve the performance of model pruning. Our method uses the regularization term to control the distribution of weights, with no need for fine-tuning after pruning if the sparsity is less than 50\%. Compared with existing fine-tuning based methods, our method can significantly improve the fine-tuned accuracy. We also explore the recovery ability of the pruned models. Our results show that the pruned models can be recovered by replacing the pruned values with their mean value. Combined with the proposed on-the-fly pruning and instant recovery, our method can generate various sparsity and accuracy models instantly.

\bibliography{aaai23}

\end{document}